\RequirePackage{fix-cm}
\documentclass[a4paper,british]{article}
\usepackage{spconf} 

\usepackage[british]{babel}

\usepackage{fancyhdr}

\usepackage[pdftex]{graphicx}
\graphicspath{{./figures/}}
\DeclareGraphicsExtensions{.pdf,.png,.jpeg,.jpg}
\usepackage{color} 

\usepackage[cmex10]{amsmath} 
\usepackage{amssymb}
\usepackage{bm}
\usepackage{esint} 
\usepackage{commath} 
\usepackage{units} 

\usepackage{algorithmicx}
\usepackage{algorithm}
\usepackage{algpseudocode}

\usepackage{url}
\usepackage{xargs} 

\usepackage{array} 
\newcolumntype{P}[1]{>{\centering\arraybackslash}p{#1}}
\newcolumntype{M}[1]{>{\centering\arraybackslash}m{#1}}

\usepackage{float} 
\usepackage[caption=false]{subfig}

\usepackage[unicode=true,%
 pdftitle={Median-Based Generation of Synthetic Speech Durations using a Non-Parametric Approach},%
 pdfauthor={Srikanth Ronanki, Oliver Watts, Simon King, and Gustav Eje Henter},%
 pdfkeywords={text-to-speech, speech synthesis, duration modelling, non-parametric models, LSTMs},%
 bookmarks=true,bookmarksnumbered=true,bookmarksopen=false,%
 breaklinks=true,pdfborder={0 0 0},backref=false,colorlinks=false,citecolor=blue]{hyperref}

\usepackage{enumitem} 

\renewcommand{\mid}{\,\ifnum\currentgrouptype=16 \middle\fi|\,}

\newcommand{\prob}{\mathbb{P}}
\newcommand{\real}{\mathbb{R}}

\newcommand{\data}{\mathcal{D}}

\DeclareMathOperator*{\argmin}{argmin}

\let\oldmarginpar\marginpar
\renewcommand\marginpar[1]{\-\oldmarginpar[\raggedleft\footnotesize #1]%
{\raggedright\footnotesize #1}}

\ninept 

\title{Median-Based Generation of Synthetic Speech Durations\\{} using a Non-Parametric Approach}
\name{Srikanth Ronanki, Oliver Watts, Simon King, and Gustav Eje Henter
}
\address{The Centre for Speech Technology Research (CSTR), The University of Edinburgh, UK\\%
  {\small \tt \href{mailto:srikanth.ronanki@ed.ac.uk}{srikanth.ronanki@ed.ac.uk}}}
\begin{document}
\ninept
\maketitle
\begin{abstract}
This paper proposes a new approach to duration modelling for statistical parametric speech synthesis in which a recurrent statistical model is trained to output a phone transition probability at each timestep (acoustic frame). Unlike conventional approaches to duration modelling -- which assume that duration distributions have a particular form (e.g., a Gaussian) and use the mean of that distribution for synthesis -- our approach can in principle model any distribution supported on the non-negative integers. Generation from this model can be performed in many ways; here we consider output generation based on the median predicted duration. The median is more typical (more probable) than the conventional mean duration, is robust to training-data irregularities, and enables incremental generation. Furthermore, a frame-level approach to duration prediction is consistent with a longer-term goal of modelling durations and acoustic features together. Results indicate that the proposed method is competitive with baseline approaches in approximating the median duration of held-out natural speech.

\end{abstract}
\section{Introduction}
\label{sec:intro}
This paper presents a new approach to the modelling and generation of speech-segment durations, a characteristic of speech which contributes to the perception of its rhythm.
Generating appropriate rhythm and melody is a challenging but vital step in producing natural-sounding synthetic speech, particularly in expressive or conversational scenarios. Steady improvements have been made in statistical parametric speech synthesis (SPSS), in particular through the adoption of deep and recurrent machine-learning techniques in recent years \cite{zen2014deep, zwu2015DNN}. As we expect that high-level, long-range dependencies are of importance for the prosodic structure of speech, recurrent models such as LSTMs are well-suited to prosodic sequence modelling problems \cite{zen2015unidirectional}.

In spite of the progress, however, the prosodic characteristics of synthetic speech remain one of its major shortcomings. One factor which we suppose contributes to this shortcoming is that current systems effectively model duration with a Gaussian distribution and generate predictions from the mean of that distribution. This is problematic, as the distribution of speech-segment durations is well known to be skewed, and so using the mean of a normal distribution fit to such observations will not produce predictions that are most typical of the process (that is, have high probability).
Another possible weakness of current approaches towards duration generation is that they typical operate as an initial stage, separate from the generation of acoustic features \cite{zen2015unidirectional}. We consider it desirable to have a single model whose parameters are learned to simultaneously generate both segment durations and the frames of acoustic features within those segments. A major motivation for this is that such a joint model would allow the simultaneous adaptation \cite{wu2015adaptation} and control \cite{watts2015sentence-level} of rhythmic, melodic and phonetic characteristics in a stable and consistent way. However, one obvious difficulty in implementing such a model is that predictions of segment durations and of acoustic observations are conventionally made on two separate time-scales. Most recently, wavenet \cite{Aaron2016wavenet} has made waveform-level modelling possible in TTS, but still requires an external duration model. Our approach can in principle be integrated with WaveNet, for sample-level duration modelling.

We here present an approach to duration modelling which in essence consists of predicting at each acoustic frame a probability that the model will subsequently advance to the next phone in the phonetic sequence. We show that -- assuming a recurrent model is used -- this approach may describe any duration distribution on the positive integers. Although it is feasible to model duration by explicitly choosing other distributions which may be more appropriate than a Gaussian (e.g., log-normal or gamma distribution), these still might not be optimal for a given conditional duration, and so the possibility of not having to commit to a predetermined distribution before model training is a powerful advantage of our approach. Furthermore, as our model moves from modelling duration at the level of the phonetic segment to the level of the acoustic frame, our approach is a necessary ingredient of our on-going work in unifying models of durations and acoustics, so that acoustic parameters a phone transition probabilities are predicted jointly for each frame.

\section{Background}
In this section, we give a brief overview of how speech-sound durations have been generated in synthetic speech, with a particular emphasis on the use of statistical methods for this task. 
This provides context for the proposed method in Section \ref{sec:proposedmethod}.

\subsection{A Brief Review of Modelling and Generating Durations}
\label{sec:durmodels}
In early, formant-based synthesis systems, phone durations were generated by rule \cite{klatt1987review}. Rules were commonly hand-crafted, rather than learned from data, meaning that no statistical modelling was used.
The concatenative synthesis approaches that emerged next did not require modelling or generating durations, since the units themselves (typically diphones) possess intrinsic durations. That said, some approaches allowed predicted durations to be incorporated into the target cost \cite{bulyko2001joint}.

The rise of statistical parametric speech synthesis (SPSS) \cite{zen2009statistical,king2011introduction} has introduced a new methodology for duration generation, in which a statistical model (probability distribution) is created to describe speech-sound durations. This distribution is supplemented by two things: a machine learning method to predict the properties of the statistical model from the input text (i.e, the linguistic features), and a principle for generating durations from the model distribution, such as random sampling or taking the mean of the distribution.

The first SPSS systems were based on simple hidden Markov models (HMMs), which implies that state durations (commonly five states per phone) are assumed follow a memoryless geometric distribution. Decision trees (DTs) were used to predict the distribution parameter based on training data, with the mean of the predicted distribution used for generation.
\begin{figure}[t]
\begin{minipage}[b]{1.0\linewidth}
  \centering
  \centerline{\includegraphics[width=0.8\columnwidth]{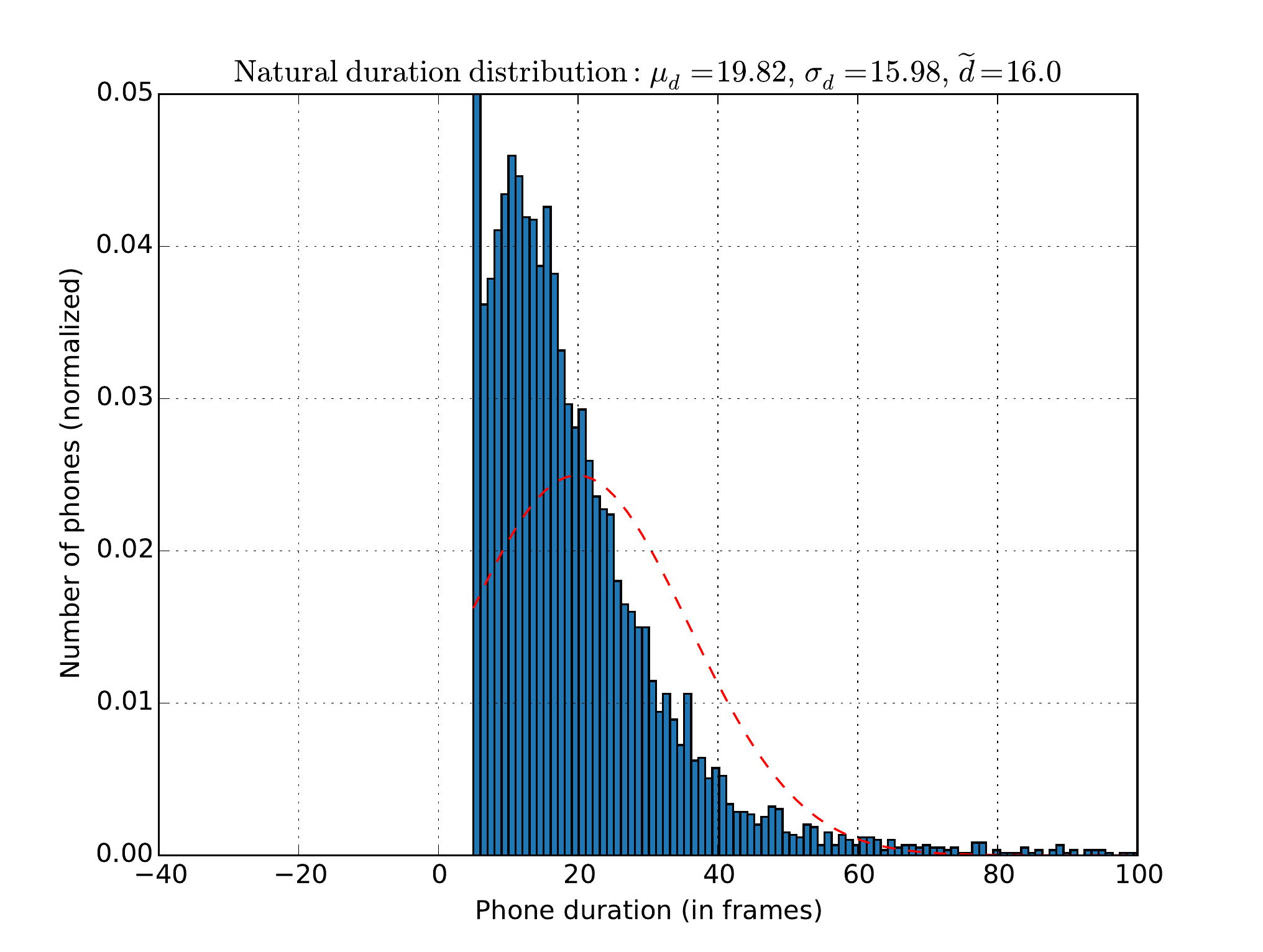}}
\caption{Distribution of all natural speech-sound durations in our held-out dataset, and a maximum likelihood-fitted Gaussian density. The median duration is 16. The minimum duration is five since the HMM-based forced aligner used five states per phone and a left-to-right topology with no skip transitions.}
\label{fig:durationhistogram}
\end{minipage}
\vspace{-1.5em} 
\end{figure}

In reality, a geometric distribution is quite far removed from how natural speech durations are distributed, as can be seen in Figure \ref{fig:durationhistogram}. Zen et al.\ \cite{zen2004hidden} introduced the idea of using hidden semi-Markov models (HSMMs) to describe durations  in the context of speech synthesis.
These models track the amount of time (acoustic frames) spent in the current HMM state, and allow the frame counter to influence the probability of transitioning, making it possible to model a much wider class of duration distributions.
Typically, HSMM durations would be assumed to follow a parametric distribution of some sort; the widely-used \emph{HMM-based speech synthesis system} (HTS) \cite{zen2007hmm} defaults to Gaussian duration distributions, for instance,  although the skewness and non-negativity of speech durations mean that other choices of distribution (log-normal, gamma) might be more suitable \cite{campbell1989syllable-level, huber1990statistical}.
As with HMMs, decision trees would be used to predict distribution parameters per state, and the mean of the predicted distribution was used for generation.

Recently, SPSS has seen stronger machine-learning techniques such as deep and/or recurrent neural networks replace decision trees for predicting the properties of duration distributions from the text, e.g., \cite{zen2015unidirectional, zen2016fast}.
Most often, these systems train DNNs or RNNs to minimise the mean squared prediction error, which is theoretically equivalent to using a Gaussian duration model (with a globally tied standard deviation) in conjunction with mean-based duration generation.
The evolution of approaches to duration generation for TTS is summarised in Table \ref{tab:duroverview}.
\begin{table}[t]
\tabcolsep=0.12cm 
\centering
\begin{tabular}{|l|c|c|c|c|}
\hline 
TTS type & Distribution & Level & Predictor & Generation \tabularnewline
\hline 
Formant & - & Phone & - & Rule\tabularnewline
Concat. & - & Phone & - & Exemplar\tabularnewline
HMM & Geometric & State & DT & Mean\tabularnewline
HSMM & Parametric & State & DT & Mean\tabularnewline
NN & Gaussian & State & DNN/RNN & Mean\tabularnewline
Proposed & Non-parametric & Frame & DNN/RNN & Median\tabularnewline
\hline 
\end{tabular}
\caption{Methods for generating speech-sound durations (last column) in different types of TTS. Statistical methods incorporate a distribution, a method of predicting its properties, in addition to a method of generating output from the predicted distribution.}
\label{tab:duroverview}
\vspace{-1.5em} 
\end{table}

Looking at Table \ref{tab:duroverview}, we see that all canonical methods predict only a few numbers per phone; at most 5 means and 5 standard deviations for Gaussian state-level predictions, of which only the means affect output generation. This renders the methods incapable of predicting general duration distributions with more degrees of freedom.
In contrast to these prior methods, we describe a fundamentally non-parametric approach in which a model is trained to predict the phone transition probability at each timestep, i.e., acoustic frame, as outlined on the final line of Table \ref{tab:duroverview}.
This set-up can in theory describe any probability mass function on the positive integers. (In practice, performance may be limited by biases in the learning algorithm used.)
It thus becomes possible to meaningfully represent any and all important properties of the duration distribution, for instance its skewness, which most models hitherto have ignored.

\subsection{Median-Based Generation}
\label{sub:mediangenbackground}
All methods in Table \ref{tab:duroverview} that predict a distribution of durations automatically support a wide variety of methods for generating output durations from this distribution. While in principle, natural speech is a random sample drawn from the true duration distribution, the output naturalness of sampling methods in speech synthesis has been found to perform poorly unless highly accurate models are used \cite{henter2014measuring}. As a consequence, most SPSS systems use deterministic generation methods, which in practice is synonymous with generating the mean duration.

We here consider a scheme where durations generated at synthesis time are based on the median -- rather than the conventional mean -- of the predicted duration distribution. Since we are no longer assuming that the duration distribution is symmetric, the median will typically differ from the distribution mean. To our knowledge, no other paper in the statistical speech synthesis literature has considered predicting a median duration different from the mean. 

Conveniently, the median can be identified from the left tail of the distribution, whereas computing the mean requires evaluating the probability mass function over its entire, infinite support. This property enables median-duration based sequential output generation with no look-ahead or other overhead, which is attractive for incremental synthesis approaches.
Unlike the mean, the median is nearly always an integer.

The advantages of generating median durations are more than computational:
For skewed distributions, including the distribution of natural speech-sound durations as graphed in Figure \ref{fig:durationhistogram}, the median is frequently closer to the peak of the distribution than the mean is; cf.\ \cite{macgillivray1981mean,basu1997mean}.
This implies that, in the spirit of most likely output parameter generation \cite{tokuda2000speech}, median-based duration prediction is likely closer to the peak density -- the ``most typical'' outcome -- than the mean is. (The mean duration is often atypically long.)
Perhaps most importantly, the median is a \emph{statistically robust} quantity, and is not affected by the tails of the real duration distribution. Statistical robustness is compelling for speech synthesis \cite{henter2016robust}, particularly for big and found datasets, as it reduces the sensitivity to errors and unexpected behaviour in the training corpus. Among other things, this could be of value with the highly expressive and variable training data used for the experiments in Section \ref{sec:set_up}.

\subsection{Frame-Level Duration Prediction}
While frame-level duration predictions are uncommon, this paper is not the first to suggest it. Watts et al.\ \cite{watts2015nst} described a joint model of duration and speech parameters, where a deep neural network was trained to simultaneously output acoustic parameters and a 5-dimensional (phone) state-duration vector for each frame. Synthesis with this type of network results in a chicken-and-egg problem, where state durations are both an input and an output of the sequence prediction network. This was solved by iteratively refining state-duration predictions over multiple passes, which is slow. In our proposed method, a single pass suffices for duration generation.
Furthermore, despite the frame-level granularity of the approach in \cite{watts2015nst}, it is not straightforward to identify a probabilistic interpretation of their scheme.

A general advantage of duration predictions made at the frame level is that they can be unified with the (traditionally distinct) prediction of acoustic features, such as pitch and vocal tract filter MGCs, that takes place for each frame. Such joint modelling of durations with acoustic properties of speech was a major factor in motivating the set-up in \cite{watts2015nst}.
It is suspected that generating durations and fundamental frequency contours that are jointly appropriate may be of importance for synthetic speech prosody; cf.\ \cite{fernandez2014prosody,ronanki2016template}.

\section{Theory}
\label{sec:proposedmethod}

\subsection{Preliminaries}
Let $p\in\{1,\,\ldots,\,P\}$ be a phone index, and let $t\in\{1,\,\ldots,\,T\}$ be an index into frames. Let further $D_p$ -- a random variable -- be the duration of phone $p$, and let $d_p\in\mathbb{Z}>0$ be an outcome of $D_p$.\footnote{In many TTS systems, $D_p$ is a vector of per-phone state durations, but we restrict ourselves to phone-level predictions in this paper; a state-based formulation is a straightforward extension.} 

Natural speech phones have different duration distributions that depend on the input text. In TTS, the properties of the distribution $D_p$ are predicted from contextual linguistic features $\bm{l}_p$ extracted from the text by the synthesiser front-end. Specifically, \emph{duration modelling} is the task of mapping the sequence of linguistic features $(\bm{l}_1,\ldots,\,\bm{l}_P)$ to a sequence of predicted duration distributions $(D_1,\ldots,\,D_P)$. \emph{Duration generation}, meanwhile, is the task of mapping $(\bm{l}_1,\ldots,\,\bm{l}_P)$ to a sequence of generated durations $(\widehat{d}_1,\ldots,\,\widehat{d}_P)$. In SPSS, one or the other of these mappings is learned from a corpus of parallel text and speech data using a machine learning method; in this paper, deep and recurrent neural networks will be used. We will write $\data$ to denote a dataset of parallel input features $\bm{l}$ and random variable outcomes $d$ used to train this predictor.

\subsection{Conventional Duration Modelling}
In statistical synthesisers that generate durations on the state or phone level, the conventional approach is to assume that the durations $D_p$ follow some parametric family $f_D$ with a parameter $\bm{\theta}\in\real^N$ ($N$ being the number of degrees of freedom), i.e.,
\begin{align}
    \prob(D_p=d) & = f_D(d;\,\bm{\theta}_p)
    \text{.}
\end{align}
Predicting the distribution $D_p$ then reduces to the stochastic regression problem of predicting the distribution parameter $\bm{\theta}_p$ from the phone-level parallel input-output dataset 
\begin{align}
    \data_p & = ((\bm{l}_1,\,\ldots,\,\bm{l}_P),\,(d_1,\,\ldots,\,d_P))
    \text{.}
\end{align}
We will write $\bm{L}_p$ to denote the linguistic information available to the predictor at $p$, which is $\bm{l}_p$ for feedforward approaches and $(\bm{l}_1,\,\ldots,\,\bm{l}_p)$ for unidirectional RNNs.

In practice, most contemporary DNN-based synthesisers do not perform full distribution modelling, but map directly from $\bm{L}_p$ to the property of the distribution $D_p$ that they wish to generate, such as its mean $\mathbb{E}(D_p)$. The dominant principle -- and the only one to be considered for the baselines in this paper -- is to tune the weights $\bm{W}$ of a DNN or RNN $d(\bm{L};\,\bm{W})$ to minimise the mean squared error (MSE) on the training dataset,
\begin{align}
    \widehat{\bm{W}}(\data_p) & = \argmin_{\bm{W}} \sum_{(\bm{L}_p,\,d_p)\in\data} (d_p-d(\bm{L}_p;\,\bm{W}))^2
    \text{;}
\end{align}
synthesis-time durations $\widehat{d}_p$ are then generated by
\begin{align}
    \widehat{d}(\bm{L}_p) & = d(\bm{L}_p;\,\widehat{\bm{W}}(\data_p))
    \text{.}
\end{align}
The theoretically optimal predictor $\widehat{d}^{\star}$ that minimises the MSE is the conditional mean, 
\begin{align}
    \widehat{d}_p^{\star}(\bm{L}_p) & = \argmin_{\widehat{d}} \mathbb{E}\left((D_p-\widehat{d})^2 \mid \bm{L}_p\right)\\
    & = \mathbb{E}(D_p \mid \bm{L}_p)
    \text{,}
\end{align}
so the end result is very similar to fitting a Gaussian duration model $f_D$ and using the mean of that fitted Gaussian to generate durations.

\subsection{Non-Parametric Duration Modelling}
We will now describe a scheme that, unlike conventional, phone-level approaches with parametric families $f_D(d;\,\bm{\theta})$, is able to model and predict arbitrary duration distributions for $D$ (restricted only by the biases of the machine learning method used). The key idea is to make predictions at the frame level about when phone transitions occur.

Assume phone durations are known up until the current frame $t$, and let $p(t')$ for $1\leq{}t'\leq{}t$ be a function that maps from 
a given frame $t' \leq t$ to the phone it has been assigned to. We can then define a frame-level sequence $\bm{L}_t$ of linguistic features
\begin{align}
    \bm{L}_t & = (\bm{l}_1,\,\ldots,\,\bm{l}_t)\\
    & = (\bm{l}_{p(1)},\,\ldots,\,\bm{l}_{p(t)})
\end{align}
up until $t$, which is constant for all frames within each phone.
For brevity, we shall write $p$ for the current phone $p(t)$. Finally, we let $t_0$ be the final frame of the previous phone, so that we can define $n_t=t-t_0\geq{}1$, the duration of the current phone so far.

We now define the \emph{transition probability} $\pi_t$ for the phone $p$ at time $t$, given the linguistic features up until this point -- that is,
\begin{align}
    \pi_t & = \prob(D_p=n_t\mid D_p\geq{}n_t,\,\bm{L}_t)
    \text{.}
\end{align}
As long as the transition probabilities satisfy $\pi_t\in[0,\,1]$ and 
\begin{align}
    \prod_{t'=t_0+1}^{\infty} (1-\pi_{t'}) & = 0
\end{align}
they induce a unique duration distribution
\begin{align}
    \prob(D_p=n_t\mid \bm{L}_t) & = \pi_t \prod_{t'=t_0+1}^{t_0+n_t-1} (1-\pi_{t'})
    \label{eq:durprob}
\end{align}
on the positive integers.

We propose to build a predictor, based on training data, that estimates $\pi_t$ from the linguistic input features. Specifically, we will train this predictor on the frame-level dataset
\begin{align}
    \data_t & = (\bm{L}_T,\,(x_1,\,\ldots,\,x_T))
    \text{,}
\end{align}
where $x_t$ is an indicator variable that equals one if and only if $t$ is the final frame of the current phone, i.e.,
\begin{align}
    x_t & = 
\begin{cases}
    1 & \text{if } t=t_0+d_p\\
    0 & \text{otherwise.}
\end{cases}
\end{align}

In this paper, we consider a deep and unidirectional recurrent neural network $x(\bm{L};\,\bm{W})$, with weights $\widehat{\bm{W}}$ trained to minimise the MSE in recursively predicting the indicator variable $x_t$ -- that is,
\begin{align}
    \widehat{\bm{W}}(\data_t) & = \argmin_{\bm{W}} \sum_t (x_t-x(\bm{L}_t;\,\bm{W}))^2
    \text{.}
\end{align}
The hypothetical predictor $\widehat{x}^{\star}$ that minimises this MSE is the conditional mean, 
\begin{align}
    \widehat{x}_t^{\star}(\bm{L}_t\,) & = \argmin_{\widehat{x}} \mathbb{E}\left((X_t-\widehat{x})^2 \mid \bm{L}_t\right)\\
    & = \mathbb{E}(X_t \mid \bm{L}_t)\\
    & = \sum_{x\in\{0,\,1\}} x\cdot\prob(X_t=x \mid \bm{L}_t)\\
    & = \prob(X_t=1 \mid \bm{L}_t)
    \text{,}
\end{align}
which is mathematically equivalent to the transition probability $\pi_t$.
This means that, as long as the predictor of $X_t$ is theoretically capable of generating arbitrary outputs for every frame, our approach can describe virtually any transition distribution -- and thus any duration distribution -- on the positive integers. LSTMs and other RNNs satisfy this requirement, due to their internal state (memory), here denoted $\bm{c}_t$, which evolves from frame to frame.

Unlike the maximum likelihood objective function, the MSE used here is not disproportionately sensitive to large errors in probability estimates of rare events (outliers). This makes the estimated $\widehat{\bm{W}}$ more statistically robust to errors in data and model assumptions.

\subsection{From Transitions to Median Durations}
\label{sub:mediangenmath}
Having defined a phone duration distribution in \eqref{eq:durprob} and trained a model to predict this distribution from data, we must consider how predicted durations $\widehat{d}$ are to be generated from this distribution. As discussed in Section \ref{sub:mediangenbackground}, sampling typically yields poor naturalness, while mean-based generation is sensitive to the tails of the distribution and unsuitable for sequential generation. On the other hand, it is straightforward to derive the right tail probability of the phone duration distribution induced by the values of $\pi_t$ seen thus far, as 
\begin{align}
    \prob(D_p>n_t\mid\bm{L}_t) & =  \prod_{t'=t_0+1}^{t_0+n_t} (1-\pi_{t'})
    \text{.}
    \label{eq:tailprob}
\end{align}
This relation straightforwardly enables synthesis based on the predicted \emph{median} duration: by stepping from $n_t=1$ and upwards, the (estimated) median duration $\widehat{d}_p$ of phone $p$ is reached when $\mathrm{P}(D>d)$ first dips down to 0.5 or below,
\begin{align}
    \widehat{d}_p & = \min_{n_t\in\mathbb{Z}} n_t\\
    & \hphantom{=} \text{such that } \prob(D_p>n_t) \leq{} 0.5 
    \text{.}
\end{align}
Expression \eqref{eq:tailprob} thus allows us to generate frames sequentially, advancing $p(t+1)$ to the next phone $p+1$ when the predicted median duration is reached, with no additional overhead at generation time.

Just as the mean squared error is minimised by the mean, the median is the theoretical minimiser of another error measure, namely the \emph{mean absolute error} (MAE),
\begin{align}
    \mathrm{MAE}(\widehat{d})& = \sum_{p\in\data_p}\left|d_p-\widehat{d}_p\right|
    \text{.}
    \label{eq:augmae}
\end{align}
If a method improves the MAE, we would expect its prediction to be closer to the (conditional) median of the data.
Interestingly, summing absolute rather than squared errors is a common component of the mel-cepstral distortion (MCD) \cite{kubichek1993mel-cepstral} often used to evaluate acoustic models in speech synthesis, but the same idea is less commonly seen for evaluating generated durations.

\subsection{Adding External Memory}\label{external_memory}
In the proposal described so far, progress through the current phone is tracked solely via the internal state of the predictor used ($\bm{c}_t$ for an LSTM). This is in contrast to the approach used by the hidden semi-Markov models in HTS, which achieved more general duration distributions than regular HMMs by maintaining an external counter of the number of frames spent in each state, and using it to compute the transition probability.
However, nothing prevents us from adding our variable $n_t$, which counts the frame number within the current phone, as an input to the neural network $x(\,\cdot\,;\,\bm{W})$ that predicts frame-level transition probabilities, in addition to the regular, linguistic features $\bm{L}_t$.
We call these augmented features $\bm{l}'_t$ and
\begin{align}
    \bm{L}'_t & = ([\bm{l}_1^{\intercal},\,n_1]^{\intercal},\,\ldots,\,[\bm{l}_t^{\intercal},\,n_t]^{\intercal})
    \text{,}
\end{align}
so that the augmented predictor is $x(\bm{L}';\,\bm{W})$.

One may surmise that having $n_t$ as an input feature could increase performance, as it provides highly relevant information to the model at all times, instead of hoping the predictor will discover the importance of duration tracking it on its own during training. To test this hypothesis, our experiments in Section \ref{sec:set_up} compare two 
systems that differ only in whether or not they incorporate an external counter $n_t$ as an input to the central neural network.

As a side note, appending $n_t$ to $\bm{l}_t$ makes the inputs to the frame-level neural network $x(\,\cdot\,;\,\bm{W})$ different with each frame. This makes it possible for the predicted transition distributions to differ from frame to frame as well, even without using a stateful predictor such as an RNN. This should, for example, enable the prediction of arbitrary transition distributions also when $x(\,\cdot\,;\,\bm{W})$ is a non-recurrent, feedforward neural network. We have, however, not explored this possibility in the current paper.

\section{Experimental set-up}
\label{sec:set_up}

\subsection{Data}
\label{ssec:data}
As a first inquiry into the viability of our frame-level duration-prediction approach, we conducted some experiments on a speech database created for to the 2016 Blizzard Challenge\cite{simon2016blizzard}.\footnote{\href{http://www.synsig.org/index.php/Blizzard_Challenge_2016}{http://www.synsig.org/index.php/Blizzard\_Challenge\_2016}} 
The database -- provided to the Challenge by
Usborne Publishing Ltd.\ -- consists of the speech and text of 50
children's audiobooks spoken by a British female speaker. We made use of a segmentation of the audiobooks carried out by another Challenge participant\footnote{Innoetics: \href{https://www.innoetics.com}{https://www.innoetics.com}} and kindly made available to other participants. The total
duration of the audio was approximately 4.33 hours after segmentation. For the purposes
of the work presented here, 4\% of the data was set aside as a test set. The
test set consists of three whole short stories: \textit{Goldilocks and
the Three Bears}, \textit{The Boy Who Cried Wolf} and \textit{The Enormous
Turnip}, having a total combined duration of approximately 10 minutes.

\subsection{Feature extraction}
We obtained a state-level forced alignment of the sentence-segmented data described above using context independent HMMs. Festvox's \texttt{ehmm} \cite{prahallad2006sub} was used to insert pauses into the annotated phone sequences based on the acoustics, to improve the accuracy of forced-aligned durations. Each phone was then characterised by a vector of 481 text-derived binary and numerical features: these features are a subset of the features used in decision-tree clustering questions from the HTS public demo \cite{zen2007hmm}; numerical features queried by those questions were used directly where possible. 
All inputs were normalised to the range $[0.01,\,0.99]$. The phone durations used to train conventional baselines was obtained by counting the frames in a phone. Contrary to our previous work \cite{henter2016robust}, sub-phone states were not used in either DNN training or prediction. This is consistent with our longer-term goal of freeing ourselves from depending on often arbitrary HMM sub-phone alignments.

\subsection{System training}
Four systems were trained: two phone-level predictors (which we identify as \textit{Phone-DNN} and \textit{Phone-LSTM}) to provide benchmarks and assess the impact of recurrent modelling on duration prediction, and two experimental systems implementing the new idea (which we identify as \textit{Frame-LSTM-I} and \textit{Frame-LSTM-E}). 

For training, all networks were initialised using small random weights,
with no pre-training. Each
duration prediction system was trained with a fixed learning rate,
manually tuned to yield close-to-optimal results on the development
set in 25 epochs or less. Early stopping was used to avoid overfitting,
by aborting training once the objective function on the development
set had failed to improve for five epochs.

\subsubsection{\textit{Phone-DNN} and \textit{Phone-LSTM}}
The two baselines use phone-level linguistic features as input and are optimised  to predict the (mean and variance normalised) duration of the phones in the training data. \textit{Phone-DNN} is a feedforward DNN with six layers of 1024 nodes each. The hidden nodes used $\mathrm{tanh}$ activation and the output was linear. \textit{Phone-LSTM} was configured with five feed-forward layers of 1024 nodes each and a final uni-directional SLSTM \cite{zwu2016investigating} hidden layer consisting of 512 nodes. 

\subsubsection{\textit{Frame-LSTM-I} and \textit{Frame-LSTM-E}}

Both proposed systems used the same architecture as that of \textit{Phone-LSTM} but -- unlike the baseline systems -- were trained with 1 datapoint per frame instead of per phone. The targets presented in training were 0.0 for non-phone-final frames, and 1.0 for phone-final frames. \textit{Frame-LSTM-I} took only the vector representing the current phone as input; at run-time it was therefore required to rely on its \textit{internal} memory (hidden state and LSTM cell state) to determine the phone transition probability at any given frame. The inputs to \textit{Frame-LSTM-I} are identical for all frames in any given phone.  \textit{Frame-LSTM-E} implemented the idea described in Section \ref{external_memory}: inputs encoding the current phone context are augmented with a frame counter indicating the number of frames that have passed since the start of the phone.  \textit{Frame-LSTM-E} can therefore also rely on this \textit{external} memory of how much time has elapsed within the current phone when making phone transition predictions, in addition to internal memory.

\subsection{Synthesis}

At synthesis time, \texttt{ehmm} phone sequences derived from the test data were used as input to each duration prediction model. This corresponds to using an oracle pausing strategy, but providing no other acoustically-derived information to the predictors. Generation from \textit{Phone-DNN} and \textit{Phone-LSTM} is straightforward: an (effectively mean) output is generated for each phone.

Median-based duration predictions were obtained from \textit{Frame-LSTM-I} and \textit{Frame-LSTM-E} by running the models recurrently over frame-level inputs and using the technique explained in Section \ref{sub:mediangenmath} and summarised in pseudocode in Algorithm 1, to decide when the median had been reached and thus when to move to the next phone. By keeping track of the time spent in each phone, we obtain median-based predictions of phone duration either with external memory of time elapsed in current phone (\textit{Frame-LSTM-E}) or without it (\textit{Frame-LSTM-I}).

\begin{algorithm}
 \caption{Switching criterion for \textit{Frame-LSTM-I}}\label{switch_prob}
 \begin{algorithmic}[1]
    \Procedure{SwitchProb}{$linguisticInput$} 
     \State $nof\gets \texttt{length of linguisticInput}$ 
     \For{each $phoneNum$ in $nof$} 
     \State $remMass \gets 1$ 
     \For{each frame $t$ in order}
        \State $ph \gets linguisticInput[phoneNum]$
        \State $probSwitchAtT \gets FrameLSTM(ph)$
        \State $remMass \gets remMass*(1-probSwitchAtT)$ 
        \If{$remMass <= 0.5$}
            \State $predictedDur[phoneNum] \gets t+1$
            \State break; 
      \EndIf
     \EndFor
     \EndFor
     \State \textbf{return} $predictedDur$ 
    \EndProcedure
 \end{algorithmic}
\end{algorithm}

\section{Results}

Table \ref{tab:objective_overall} gives RMSE, mean absolute error (MAE) and Pearson correlation computed by comparing the predicted durations against the held-out reference. Silence segments were excluded from these calculations. 

Firstly, results for the benchmark systems show that \textit{Phone-LSTM} outperforms \textit{Phone-DNN} in every respect, confirming the importance of the LSTM's recurrence for duration modelling. Turning to the experimental systems, it can be seen that while the mean-based predictions of the LSTM and the DNN baselines clearly outperform the proposed methods in terms of RMSE, the gap is much smaller when it comes to MAE. Median-based generation methods are expected to score better on MAE than RMSE, because -- as discussed in Section \ref{sub:mediangenmath} -- the median is the theoretical minimiser of MAE, while the mean minimises the (R)MSE. 

The breakdown of results by phonetic class given in Table \ref{tab:objective_by_phoneclass} shows an interesting phenomenon: while the proposed system \textit{Frame-LSTM-E}'s MAE is worse than the LSTM benchmark for vowels and slightly worse for consonants overall, for all classes of consonant except plosives the proposed method performs better. 

While the best-performing experimental system does not overall outperform the LSTM baseline even on MAE, the gap in performance is effectively closed (4.556 vs.\ 4.574). This means that we have devised a system which gives similar performance in the MAE sense as our existing one, but provides greater compatibility with our acoustic models, as it too operates at the frame level. The next logical step, therefore, is to apply this idea to joint modelling of duration and acoustic features.

\begin{table}[t]
\tabcolsep=0.12cm 
\centering
\begin{tabular}{|l|c|c|c|}
\hline
\textbf{Model} & \textbf{RMSE}  &  \textbf{MAE} &  \textbf{Corr.}  \\
\hline
Phone-DNN    &  8.037 & 4.759 &  0.750       \\
Phone-LSTM   &  7.789 & \textbf{4.556} &  0.765       \\
\hline
Frame-LSTM-I &  8.254 & 4.610  &  0.761      \\
Frame-LSTM-E &  8.294 & \textbf{4.574}  &  0.754      \\
\hline
\end{tabular}
\caption{RMSE and MAE (both in frames per phone), along with Pearson correlation of the predicted durations, all measured w.r.t.\ force-aligned durations.}
\label{tab:objective_overall}
\end{table}

\begin{table}[t]
\tabcolsep=0.12cm 
\centering
\begin{tabular}{|l|c|c|c|c|c|c|c|}
\hline
 & \multicolumn{3}{c|}{\textbf{Phone-LSTM}} & \multicolumn{3}{c|}{\textbf{Frame-LSTM-E}} \\
\cline{2-7}
\multicolumn{1}{|l|}{\textbf{Phonetic class}} &    \textbf{RMSE}     &  \textbf{MAE} &  \textbf{Corr.} & \textbf{RMSE} & \textbf{MAE}  & \textbf{Corr.} \\
\hline
Vowels &   8.516  & \textbf{4.848} &  0.809   &  9.027 & 4.891 &  0.799       \\
Consonants &   7.313  & \textbf{4.378} &  0.709   &  7.815 & 4.382 &  0.694        \\
\hline
Plosives &  5.206 & \textbf{3.608} &  0.732   & 5.610 & 3.612  &  0.720       \\
Fricatives &   6.489 & 4.380 &  0.769  & 6.859 & \textbf{4.246}  &  0.764      \\
Nasals &   6.833 & 4.459 &  0.568  & 7.376 & \textbf{4.398}  &  0.550      \\
Affricates  &   5.658 & 4.220 &  0.797  & 5.432 & \textbf{3.746}  &  0.821      \\
Glides + liquids &  8.013 & 5.260 &  0.569  & 8.235 & \textbf{5.075}  &  0.599      \\
\hline
\end{tabular}
\caption{Objective measures from Table \ref{tab:objective_overall} broken down by  phonetic class.}
\label{tab:objective_by_phoneclass}
\vspace{-1.5em} 
\end{table}

\section{Conclusions and Future Work}

We have described a new duration modeling paradigm with LSTMs which operates at frame-level for duration prediction in speech synthesis. Experiments conducted on an audiobook data were found to perform competitively when compared with a baseline phone-level LSTM system. As a next step, we will explore the generation based on other quantiles such as mean so as to allow the change in rate of speaking. Also, the frame-level joint acoustic/duration model should allow iterative refinement of the data alignments. Future work includes joint modeling of duration and acoustic features along with subjective evaluation of synthesised speech.

\section{Acknowledgements}
This research was supported by EPSRC Programme Grant EP/I031022/1, Natural Speech Technology (NST).
The NST research data collection may be accessed at \url{http://hdl.handle.net/10283/786}.

\bibliographystyle{IEEEbib}
\bibliography{refs}

\end{document}